\pdfoutput=1

\documentclass[11pt]{article}

\usepackage[]{emnlp2021}

\usepackage{times}
\usepackage{latexsym}

\usepackage[T1]{fontenc}

\usepackage[utf8]{inputenc}
\usepackage{graphicx}
\usepackage{microtype}
\usepackage{color}
\usepackage{soul}
\usepackage{multirow}
\usepackage{makecell}
\usepackage{amssymb}
\usepackage{amsmath}
\usepackage{enumitem}
\usepackage{booktabs}
\usepackage{soul}
\usepackage{color}
\usepackage{array}
\usepackage{pifont}
\usepackage{arydshln} 
\usepackage{makecell}
\newcommand{\red}[1]{{\color{red}{#1}}}
\newcommand{\orange}[1]{{\color{orange}{#1}}}
\newcommand{\blue}[1]{{\color{blue}{#1}}}

\usepackage{titlesec}
\titlespacing*{\section}{0pt}{*0.8}{*0.8}
\titlespacing*{\subsection}{0pt}{*0.6}{*0.6}

\newcommand{\Skip}[1]{}

\newcommand{\ie}{\textit{i}.\textit{e}.\ }

\newcommand{\secref}[1]{Section \ref{#1}}
\newcommand{\figref}[1]{Figure \ref{#1}}

\newcommand{\dotieconcat}[2]{
  \text{\raisebox{.8ex}{$\smallfrown$}}%
}

\title{Character-Centric Story Visualization via \\Visual Planning and Token Alignment}

\author{Hong Chen$^{1}$\thanks{~Work done when the author was visiting UCLA.}, Rujun Han$^{3}$, Te-Lin Wu$^{2}$, Hideki Nakayama$^{1}$ and Nanyun Peng$^{2}$
\\ The University of Tokyo$^1$, University of California, Los Angeles$^2$, AWS AI Labs$^3$  \\  \texttt{\{chen, nakayama\}@nlab.ci.i.u-tokyo.ac.jp}\\\texttt{rujunh@amazon.com}, \texttt{\{telinwu,violetpeng\}@cs.ucla.edu}
}

\begin{document}
\maketitle
\begin{abstract}
\textbf{Story visualization} advances the traditional text-to-image generation by enabling multiple image generation based on a complete story. This task requires machines to 1) understand long text inputs and 2) produce a globally consistent image sequence that illustrates the contents of the story. A key challenge of consistent story visualization is to preserve \textit{characters} that are essential in stories. To tackle the challenge, we propose to adapt a recent work that augments Vector-Quantized Variational Autoencoders (VQ-VAE) with a text-to-visual-token (transformer) architecture. Specifically, we modify the text-to-visual-token module with a two-stage framework: 1) \textbf{character token planning} model that predicts the visual tokens for \textit{characters only}; 2) \textbf{visual token completion} model that generates the remaining visual token sequence, which is sent to VQ-VAE for finalizing image generations. To encourage characters to appear in the images, we further train the two-stage framework with a character-token alignment objective. Extensive experiments and evaluations demonstrate that the proposed method excels at preserving characters and can produce higher quality image sequences compared with the strong baselines. Code can be found in ~\url{https://github.com/PlusLabNLP/VP-CSV}

\end{abstract}
\section{Introduction}

\begin{figure}[t]
    \centering
    \includegraphics[scale=0.35]{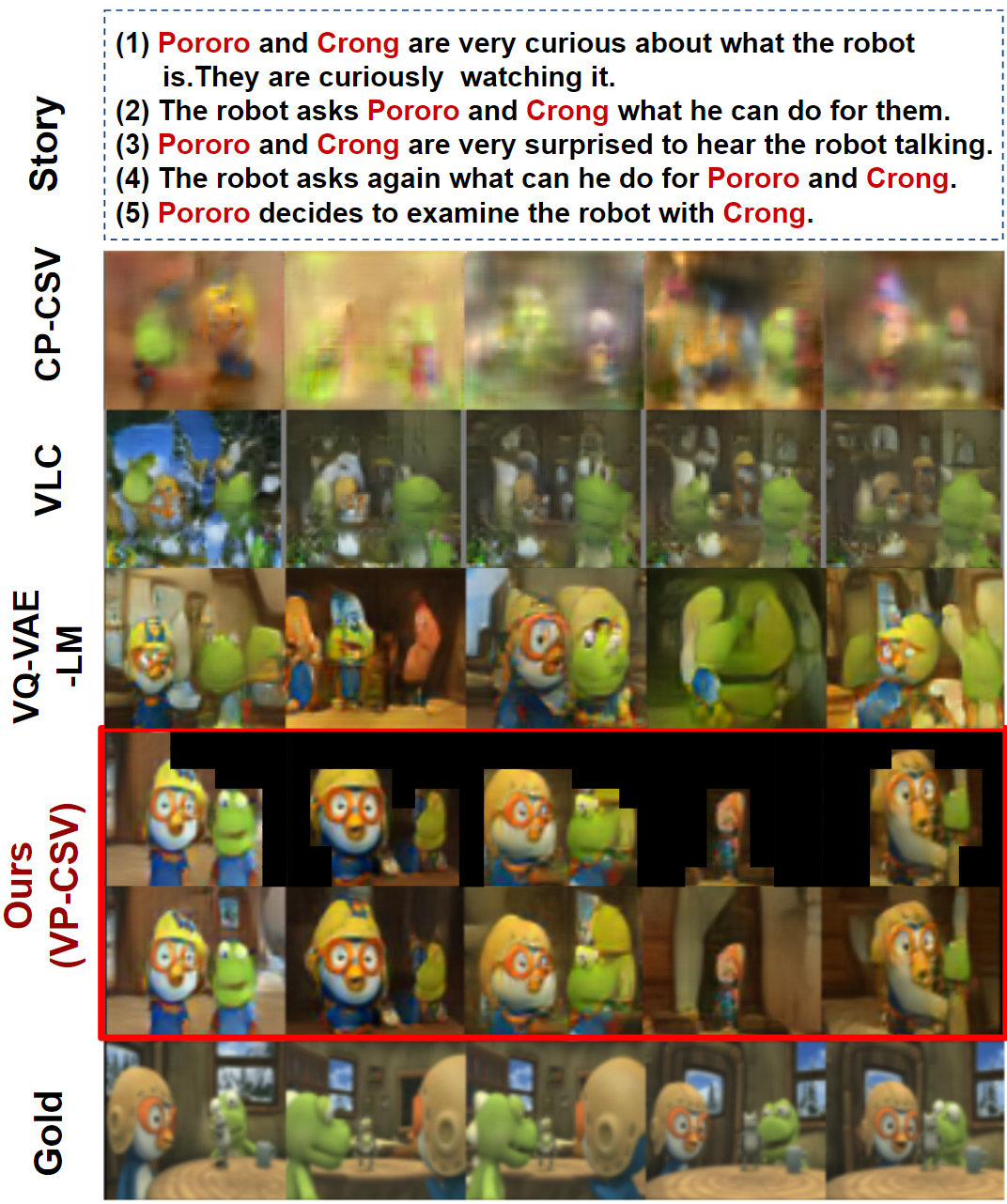}
    \vspace{-0.2cm}
    \caption{Example of generated images from previous models (CP-CSV and VLC), the VQ-VAE with transformers (VQ-VAE-LM) baseline, and our proposed model (VP-CSV). We highlight the characters in each sentence in red. We can see that the generated image sequence by VP-CSV achieves the best image quality and character preservation.}
    \label{fig:first}
    \vspace{-0.5cm}
\end{figure}

Text-to-image generation~\cite{ramesh2021zero, ding2021cogview, qiao2019mirrorgan}, as a benchmark task to test AI systems' multi-modal capability, has been widely studied over the past decade. The task takes a short sentence or phrase as input and outputs a single image illustrating the content of the input text. One of the latest successful works is \textsc{DALL-E} \citep{ramesh2021zero}, which efficiently encodes and decodes images with discrete visual tokens and achieves remarkable text-to-image generation performance by prefixing visual tokens with natural language inputs.

Recently, the \textbf{story visualization} task has attracted increasing research interests with a variety of potential and promising applications such as automatic production of movie clips and animations from written scripts. Story visualization is more challenging than the original text-to-image generation as it requires models to generate a sequence of images that visually illustrate a story composed of at least several sentences. Machines need to understand long story texts and simultaneously generate images with consistent scenes and characters described in the story coherently. 

Character is one of the essential elements of a story since they lead and develop the story~\cite{montgomery2004language}. Further, since characters usually occupy a large region in an image, incorrect or vague characters would result in low-quality image generation. Therefore, ensuring character preservation is a key cornerstone for consistent story visualization, which motivates the character-centric approach in this work. Our character-centric method is based on a recent state-of-the-art text-to-image generation model, VQ-VAE-LM that combines VQ-VAE~\citep{van2017neural} with a text-to-visual-token transformer \citep{ramesh2021zero, yan2021videogpt}. In order to boost character consistency in story visualization, we propose to enhance VQ-VAE-LM with a two-stage module inspired by Plan-and-Write story generation framework~\citep{peng2018towards, yao2019plan, goldfarb2019plan, goldfarb-tarrant-etal-2020-content, han2022go}. We briefly summarize the approach below.

The two-stage model freezes VQ-VAE and adapts the text-to-visual-token transformer by dividing it into two separate modules: 1) \textbf{the plan module} generates a token plan consisting of character and non-character (background) tokens, which reinforces our system's attention to characters; 2) based on the character token plan, \textbf{the completion module} produces the entire sequence of visual tokens, which are used in VQ-VAE decoder to produce final images. The two modules are separately trained to avoid the expensive decoding in training. Crucially, we train \textbf{the completion module} with an auxiliary semantic alignment loss that encourages appropriate character tokens to appear in the final visual tokens.

To the best of our knowledge, all previous works in story visualization employ GAN-based~\cite{goodfellow2014generative} methods, which are shown to be unstable~\cite{kodali2017convergence,thanh2018improving}. Figure~\ref{fig:first} shows that the images generated by VQ-VAE-LM are clearer than those by previous GAN-based models such as CP-CSV~\cite{song2020character} and VLC~\cite{maharana2021integrating}, suggesting the instability of GAN-based models. On the other hand, VQ-VAE-LM architecture has been shown to stabilize the training ~\citep{van2017neural}, leading to better image quality (the third row in Figure~\ref{fig:first}). Finally, our proposed two-stage visual planning model (VP-CSV) can further improve VQ-VAE-LM, which produces the best quality images in Figure~\ref{fig:first}.

Beyond this example, we conduct extensive experiments to show that our proposed two-stage visual token plan and the semantic alignment method perform better on character preservation compared with various baseline models, which in general contribute to the improved story visualization quality. 
We summarize our contributions below,
\begin{itemize}[noitemsep]
    \item We adopt the VQ-VAE-LM architecture as our baselines for the story visualization task, which achieve better results than previous GAN-based models.
    \item Our key contribution is to develop VQ-VAE-LM with visual token planning and alignment to improve the performance of character preservation and visual quality.
    \item Extensive experiments, evaluations and analysis show that our models outperform all baseline systems by a large margin, demonstrating the efficacy of our proposed method.
\end{itemize}

\section{Related work}

\subsection{Text-to-image Generation}
In the field of image generation, Generative Adversarial Nets (GAN)~\cite{goodfellow2014generative}, Variational Autoencoders (VAE)~\cite{kingma2014auto} and their variants~\cite{yu2019vaegan,mirza2014conditional,sohn2015learning} have been studied. Text-to-image generation includes text conditions into these models. 
StackGAN~\citep{zhang2017stackgan} proposes a sketch-refinement process that first sketches the primitive shape and colors before finalizing the details.
AttnGAN ~\cite{xu2018attngan} allows attention-driven, multi-stage refinements of fine-grained text-to-image generation to improve the image quality.
MirrorGAN~\cite{qiao2019mirrorgan} further enhances the semantic consistency between the text description and visual content by leveraging semantic embedding, global attention, and semantic alignment modules.
More recently, VQ-VAE with transformer (VQ-VAE-LM) structure significantly outperforms GAN-based methods, and DALL-E~\cite{ramesh2021zero} is one of the most well-known methods that use this structure. 

\subsection{Story Visualization}
\citet{li2019storygan} is one of the earliest works to propose a Sequential Conditional GAN as a method for story visualization. They encode the story and decode the images using a sequential model, combining image and story discriminators for adversarial learning. 
\citet{maharana2021improving} extends the GAN structure by including a dual learning framework that uses video captioning to reinforce the semantic alignment between the story and generated images. Based on this new framework, ~\citet{maharana2021integrating} improves the generation quality by incorporating constituency parse trees, commonsense knowledge, and visual structure via bounding boxes and dense captioning.
To improve the global consistency across dynamic scenes and characters in the story flow, \citet{zeng2019pororogan} enhances the sentence-to-image and word-to-image-patch alignment by proposing an aligned sentence encoder and attentional word encoder.
With similar motivation, \citet{li2020improved} includes dilated convolution in the discriminators to expand the receptive field and weighted activation degree to provide a robust evaluation between images and stories.

To preserve character information, CP-CSV~\cite{song2020character} adapts a GAN-based model with figure-background segmentation that creates masks to preserve characters and improve story consistency. Our work differs from it by proposing a two-stage visual planning model enhanced by character-token alignments that can be easily incorporated into the state-of-the-art VQ-VAE-LM architecture.


\section{VQ-VAE-LM}
In this section, we review the VQ-VAE architecture and how it can be extended to a text-to-image variant, dubbed VQ-VAE-LM. We refer readers to the original VQ-VAE paper for more details.

\begin{figure}[t!]
    \centering
    \includegraphics[width=0.97\columnwidth]{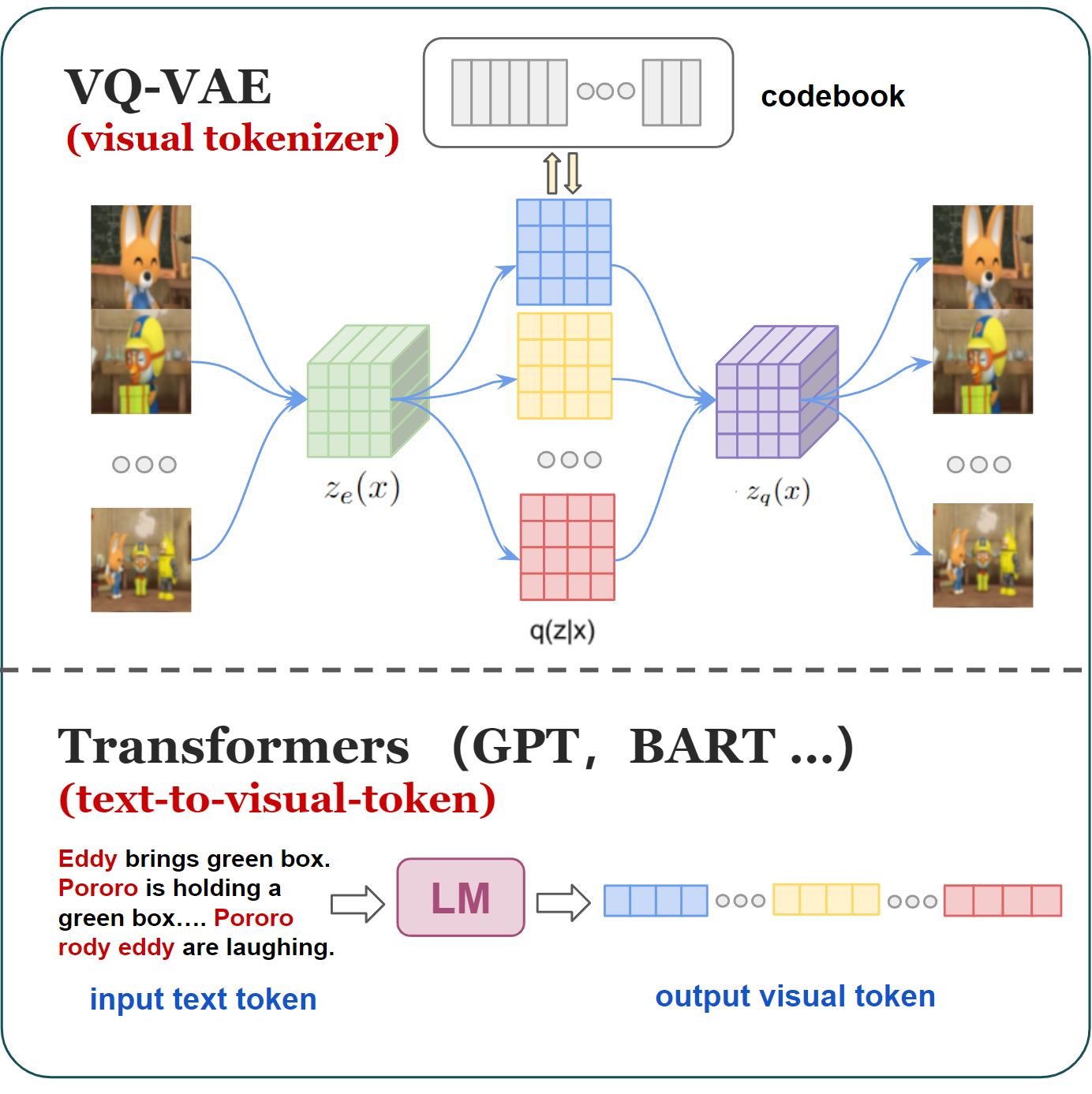}
    \caption{Overview for the VQ-VAE-LM architecture. VQ-VAE model produces visual tokenizers which are used to train the language model (transformer).
    }
    \label{fig:vqvae}
    \vspace{-0.3cm}
\end{figure}

\begin{figure*}[h!]
    \centering
    \includegraphics[scale=0.38]{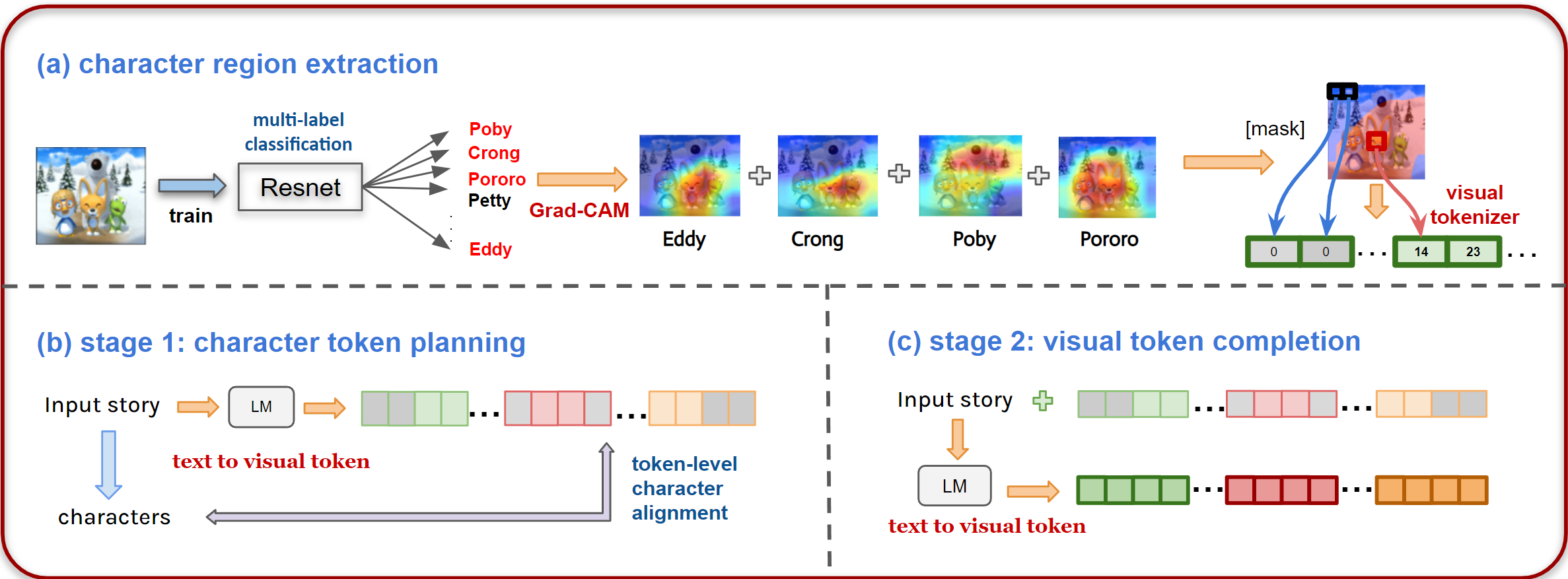}
\caption{
Pipeline of visual planning. \textbf{(a)} We first extract the character regions from the image. Based on the corresponded sentence(s), we can get the characters ``Poby'', ``Crong'', ``Pororo'' and ``Eddy'' that are involved in the image. We then apply Grad-CAM on a pretrained ResNet by specifying each character and obtaining a heatmap for each character. Combining these heatmaps, we can obtain an overall heatmap that distinguishes the character and background regions. We mask the background tokens and leave the rest character visual tokens as targets in our proposed visual planning methods. \textbf{(b)} and \textbf{(c)} depict the two-stage generation, where (b) learns to generate character visual tokens which is generated from (a), and then (c) completes the visual tokens.
}
\label{fig:pipeline}
\vspace{-0.2cm}
\end{figure*}

\subsection{VQ-VAE}
Vector Quantised-Variational AutoEncoder (VQ-VAE)~\cite{van2017neural} encodes images into latent discrete codes instead of continuous variables as in VAEs~\cite{kingma2014auto}, which helps alleviate the ``posterior collapse'' issue in VAEs whose continuous latent variables tend to be ignored by strong autoregressive decoder.

Denote the input image as $x$ and the latent variable as $z$, and both the prior $p(z)$ and the posterior $q(z|x)$ in VQ-VAE follow categorical distributions.
Assuming there are $K$ quantized entries, each $z_k$, $k\in\{1...K\}$ is associated with a trainable embedding vector $e_k$.
The encoder maps the input to a vector $z_e(x)$ that derives the posterior distribution:
\begin{equation}
\small
q(z=k \mid x)= \begin{cases}1 & \mathrm{k}=\operatorname{argmin}_{j}\left\|z_{e}(x)-e_{j}\right\|_{2} \\ 0 & \text {otherwise }\end{cases}
\end{equation}
Intuitively, the nearest embedding vector, $\hat{e_k}$, to $z_{e}(x)$, is used as the latent representation $z$. The decoder then takes as input the $\hat{e_k}$ to approximate the distribution $p(x|z)$ over the data.
The training objective of VQ-VAE is:
\begin{equation}
\small
\mathcal{L}=\underbrace{\|x-z_q(x)\|_{2}^{2}}_{\mathcal{L}_{\text {recon }}}+\underbrace{\|sg[z_e(x)]-e\|_{2}^{2}}_{\mathcal{L}_{\text {codebook }}}+\underbrace{\beta\|s g[e]-z_e(x)\|_{2}^{2}}_{\mathcal{L}_{\text {commit }}}
\end{equation}
where $sg$ denotes \textit{stop-gradient} during training.
The above objective consists of:
(1) A reconstruction loss $L_{recon}$ that encourages the model to learn good representations that accurately reconstruct the original input image $x$.
(2) A codebook loss $L_{codebook}$ that minimizes the distance between codebook embeddings ($e_k$) and the corresponding encoder outputs $z_e(x)$.
(3) A commitment loss $L_{commit}$ (usually weighted by a hyperparameter $\beta$) which prevents the encoder outputs from fluctuating too much under different code vectors.

\subsection{Text-to-Visual-Token Transformer}
\label{sec:text_to_visual_token}

To extend the standard VQ-VAE, which originally takes inputs and outputs of the same modality, to handle multimodal inputs and outputs such as text-to-image generations, the key idea is to augment the VAE architecture with a transformer-based language model (LM).
Denote the textual inputs as $y$, the text-to-image generative process can be decomposed into $p(x|y) = p(x|z)p(z|y)$.
The LM ($p(z|y)$) generates latent codes $z$ conditioned on $y$, which can be directly plugged into the decoder of VQ-VAE to reconstruct the output image.
In the existing VQ-VAE-LM architectures such as~\citet{yan2021videogpt, ramesh2021zero}, the LM component is separately trained with a standard MLE objective that maximizes the likelihood of natural sequences. 

In order to exploit the trained VQ-VAE decoder, the output space of the LM needs to be bridged to the input space of the decoder, \ie we require the LM to be finetuned to produce plausible \textit{visual tokens}.
As shown in the upper half of Figure~\ref{fig:vqvae}, the visual tokens are quantized latent codes encoded from the target image sequence.
They are then flattened and concatenated into one long token sequence.
The LM (Figure~\ref{fig:vqvae} lower half) takes these visual tokens as the training targets for encoding input stories.
We follow the existing work~\cite{yan2021videogpt, ramesh2021zero} to train VQ-VAE and LM independently; however, both components are trained \textit{from scratch} to accommodate the targeted animation domain in this work.


\section{VP-CSV: Visual Planning based Character-centric Story Visualization}
The proposed VP-CSV framework is illustrated in~\figref{fig:pipeline}.
In order to exploit character-centric inductive biases to guide a more consistent story visualization, we decompose the text-to-visual-token generation (\secref{sec:text_to_visual_token}) into a two-stage framework comprising:
(1) a~\textbf{plan module} (Figure~\ref{fig:pipeline}b) that performs character-level planning by focusing on generating visual tokens for~\textit{character} regions, while \textit{non-character} regions are masked-out, given the input story, and
(2) a~\textbf{completion module} (Figure~\ref{fig:pipeline}c) which conditions on both the input story and the outputs from (1) to produce the complete visual tokens as the input to the VQ-VAE decoder.
We will first define the notations used throughout the paper and discuss each module in detail.

\subsection{Notations}
Given a story $s$ which consists of a sequence of short paragraphs: $s=\{\mathrm{s}_1, \mathrm{s}_2, ... ,\mathrm{s}_n\}$. Each paragraph (one or a few sentences) corresponds to an image in the sequence.
The goal of story visualization is to generate a sequence of images $x=\{\mathrm{x}_1, \mathrm{x}_2, ... ,\mathrm{x}_n\}$ conditioned on the story $s$.
As discussed in~\secref{sec:text_to_visual_token}, the target images are transformed into visual tokens $z=\{\mathrm{z}_1, \mathrm{z}_2, ... ,\mathrm{z}_n\}$, where $\mathrm{z}_i \in (p \times p)$ and $p$ denotes the number of patches along each image dimension.
We flatten $\mathrm{z}_i^{p \times p}$ into [$\mathrm{z}_i^{0,1} ... \mathrm{z}_i^{0,p-1};\mathrm{z}_i^{1,0} ... \mathrm{z}_i^{p-1, p-1}$] to perform sequence generation.
For visual tokens $\mathrm{z}_i$ in the $i$-th image, we use $\mathrm{r}_i \in (p \times p)$ to denote the tokens within the regions containing story characters, while tokens in non-character regions will be replaced with a special masking token \texttt{[MASK]} as the target of our planning module.
$\mathrm{z}_i$ would be completely masked out (filled with all \texttt{[MASK]}) if the image $\mathrm{x}_i$ does not contain any characters.

\subsection{Visual planning (VP)}
\label{sec:vp}

The visual planning stage focuses on generating a visually and story-wise consistent character plan to determine the suitable spatial regions in each image frame to place characters before filling in other visual details. 
Due to the lack of annotations for character regions in the dataset, we first leverage external tools to extract character regions during data pre-processing.


\vspace{.2em}

\noindent \textbf{Character region extraction.}
As shown in Figure~\ref{fig:pipeline}a, we first train a multi-label classifer that is able to identify all characters in text inputs. We then utilize
Grad-CAM~\cite{selvaraju2017grad}, a widely-used technique for generating class-specific heatmaps to locate highly-attended regions within images.
In our case, these classes refer to characters, so Grad-CAM can help us find the character regions without any supervised data. 
For each character identified by the multi-label classifer in the associated sentence(s) of an image, Grad-CAM produces a heatmap highlighting the character region. We merge these heatmaps into one that represents the region covering all existing characters.

More specifically, we first train a CNN character classification model\footnote{The trained model achieves 86\% classification accuracy.} with multi-label classification loss, using the \textit{character mentions} in the story sentences as target labels.
As exemplified in Figure~\ref{fig:pipeline}a, we obtain four independent heatmaps for their corresponding characters and merge them into a single one.
A visual token belongs to the background (\ie the blue image patches) if none of the above character heatmaps attend to its corresponding image region.
Formally, the masks separating the character and non-character visual tokens are derived using the following formula:
\begin{equation}
\begin{split}
    \mathrm{a}^{p\times p} & = \max(\mathrm{a}_{c_1}, \mathrm{a}_{c_2}, \dots, \mathrm{a}_{c_N}),\\
    \mathrm{r}^{j,k} & = \left\{\begin{matrix}
     \mathrm{z}^{j,k} &  \mathrm{a}^{j,k} >= \gamma &  \\
     0 &  \mathrm{a}^{j,k} < \gamma\\
    \end{matrix}\right.
\end{split}
\end{equation}
where $\mathrm{a}_{c_n}$ denotes the heatmap $\mathrm{a}$ for the $n$-th identified character $c_n$. $\gamma$ is the threshold that separates the character regions from the background (masks).

\vspace{.2em}

\noindent \textbf{Character token planning (first stage).}
We adopt a GPT-2~\cite{radford2019language} language model in the \textbf{plan module} to produce (plan) the character visual tokens.
Specifically, conditioning on the input story sentences $s = \{\mathrm{s}_1,\mathrm{s}_2,...,\mathrm{s}_n\}$, the model learns to generate the planned character visual tokens $r = \{\mathrm{r}_1,\mathrm{r}_2,...,\mathrm{r}_n\}$ that are prepared in the aforementioned character-region extraction stage. 
Denote $\theta$ as the parameters of the GPT-2 model, per sample training loss can be computed as: $$L_{\theta} = - \log p (r|s, \theta)$$

\noindent \textbf{Visual token completion (second stage).}
To reinforce the model's attention on the produced character visual tokens, we use the same GPT-2 with shared parameters from the previous stage.
In this completion stage, the model is trained to generate the comprehensive (background infilled) visual tokens, conditioning on both the input stories and the character visual tokens from the first stage. 
Per sample training loss is derived as: $$L_{\theta} = - \log p (z|s, r, \theta)$$
These visual tokens will later be decoded by the trained VQ-VAE decoder model for image reconstruction.

\begin{figure}
    \centering
    \includegraphics[scale=0.3]{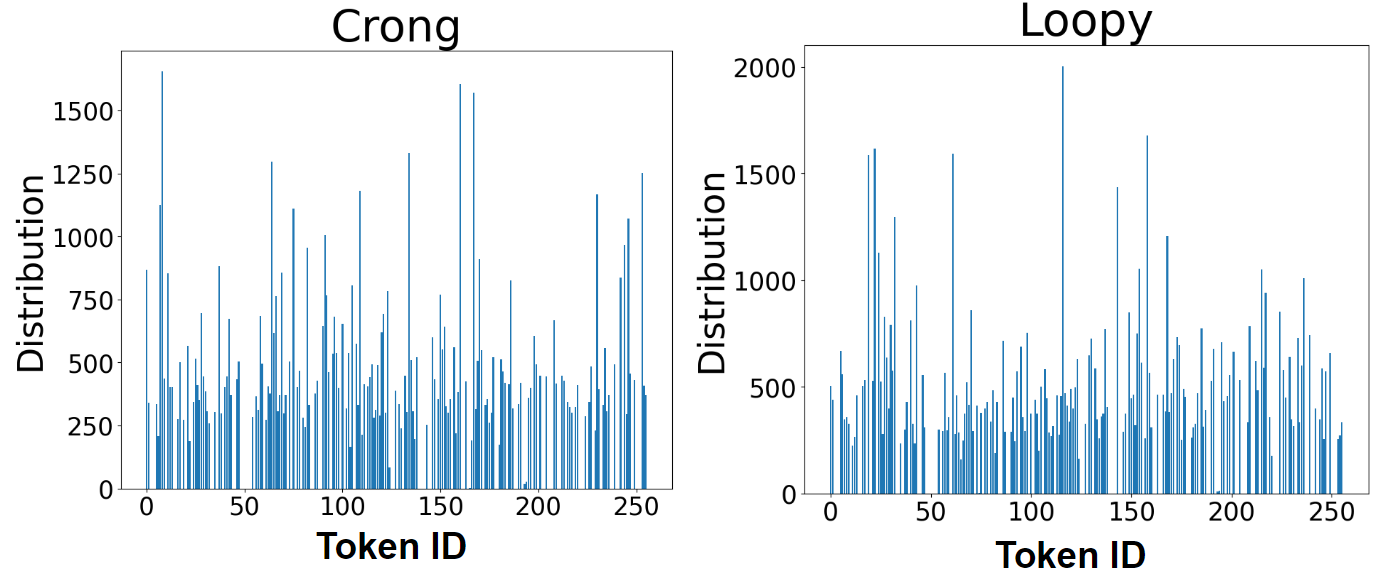}
    \caption{Token occurrence distribution per character. The top-5 frequent tokens for ``Crong'' and ``Loopy'' are [8, 160, 167, 134, 64] and [116, 158,  22,  61,  19].}
    \label{fig:tokendis}
    \vspace{-0.4cm}
\end{figure}

\subsection{Token-level character alignment (TA)}
\label{sec:token}

The data pre-processing described in the previous section enables us to obtain character visual tokens in the training data. With that, we can compute the visual token distribution for each character.
It can be observed from the token distribution shown in Figure~\ref{fig:tokendis}, that different characters tend to reflect different representative token IDs.
Based on this observation, a straightforward way to improve the model's character preservation is to increase the occurrence probability of a character's top visual tokens if it is mentioned in the story sentences.
In light of this, we propose to use a semantic loss~\cite{xu2018semantic} to encourage the character-to-visual-token alignment.

We extract the top-10 frequent tokens $t_c$ for each character $c$, and for each input story $s$, we compose a token set $T = \bigcup\limits_c t_c$, in which the character $c$ is mentioned at least once in story $\boldsymbol{s}$.
We denote the visual tokens extracted from their corresponding images as $\boldsymbol{z}$, and write the semantic loss as:

{
\begin{align} \label{eq:semantic-Obj}
\mathcal{L}^s(Q, \boldsymbol{p}) = & -\log \sum_{\boldsymbol{z} \models Q} \prod_{\boldsymbol{z}^j  \in P} p_j \prod_{\boldsymbol{z}^j \in N} (1 - p_j)
\nonumber\end{align}
}where $Q$ is the character token constraint that requires \textbf{the character tokens in $T$ to appear in the generated visual token sequence}.
$P$ indicates the token set $\{\boldsymbol{z}^j \in T \mid \boldsymbol{z}^j \in \boldsymbol{z} \}$;
$N$ denotes the token set $\{\boldsymbol{z}^j \notin T \mid \boldsymbol{z}^j \in \boldsymbol{z} \}$, and $p_j$ is the predicted probability of $\boldsymbol{z}^j$.
The intuition of this objective is that if all (identified) characters' top visual tokens show up in the predicted $\boldsymbol{z}$ (\ie $\boldsymbol{z} \models Q$), we further increase the probability of tokens in $P$, while penalizing the likelihood of generating less relevant tokens belonging to $N$. 
\section{Experimental Setup}

\begin{figure}
    \centering
    \includegraphics[scale=0.5]{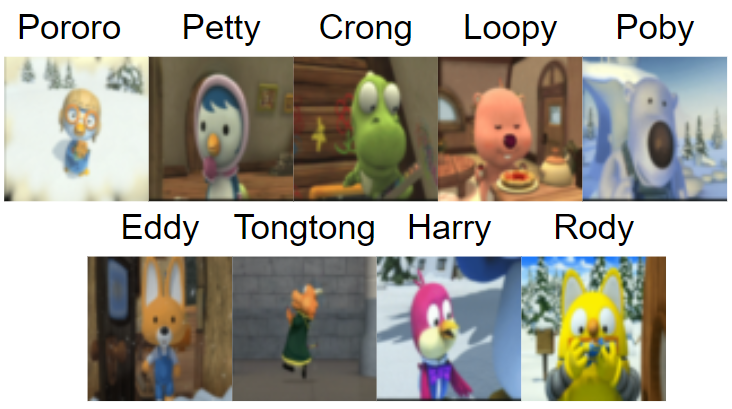}
    \caption{Main characters in Pororo-SV dataset.}
    \label{fig:character}
    \vspace{-0.6cm}
\end{figure}

\subsection{Dataset}
In this work, we use a popular story visualization dataset, Pororo-SV~\cite{li2019storygan} to train and evaluate our models.
Each story in Pororo-SV contains 5 short paragraphs (a paragraph typically consists of 1-2 sentences). Each short paragraph corresponds to one image in the target image sequence.
The dataset contains 10,191, 2,334, and 2,208 samples in train, validation, and test split. The validation split may include frames in the training split, while the test split contains all new frames \textit{unseen} from the training data. To better demonstrate the visualization performance of our model, we mainly experiment on the test split.
Besides, to help the reader understand our examples, we show the main characters of this dataset in Figure~\ref{fig:character}. 

Note that the Pororo-SV dataset is the only one we are aware of that has been studied with systematic quantitative evaluations. Another notable story visualization dataset is Flinstones~\cite{gupta2018imagine}, however, there is not established automatic evaluation for this dataset. 
To show the generalization of our proposed VP-CSV method, we put the generation example of our model trained on the Flintstones dataset in Appendix Figure~\ref{fig:flintstone}.

\subsection{Evaluation Metrics}
We adopt the \textbf{automatic evaluation metrics} following existing works of the story visualization task and report results using the evaluation script provided in prior work.\footnote{https://github.com/adymaharana/VLCStoryGan}

\noindent\textbf{Character F1 score} measures the percentages of characters obtained from the generated images that exactly appear in the story inputs. A pretrained inception v3 model~\cite{szegedy2016rethinking} is finetuned on Pororo-SV by a multi-label classification loss and can predict characters in test images.

\noindent\textbf{Frame Accuracy (Exact Match)} metric measures whether all characters in a story are present in the corresponding images using the same model as in the Character F1 score. 
Character F1 score calculates the proportion of characters existing in a story, while Frame Accuracy calculates the percentage of samples involving all characters.

\noindent\textbf{Frechet Inception Distance (FID)} summarizes how similar two groups are in terms of statistics on vision features of the raw images calculated using the same inception v3 model. Lower scores indicate that the predicted images are more similar to the ground-truth images.

\noindent\textbf{Video Captioning Accuracy} conducts back translation (\ie image-to-text generation) to measure the global semantic alignment between captions and generated visualizations. This metric utilise a video captioning model that is trained with the Pororo-SV dataset to generate stories based on the image sequences. It reports BLEU scores between the predicted caption and the ground truths (the story inputs).

\noindent\textbf{R-Precision~\cite{xu2018attngan}} is a retrieval-based metric that computes the probability that a generated image sequence has the highest semantic similarity with its input story among all stories in the test data.
a Hierarchical DAMSM~\cite{maharana2021improving} is used to compute cosine similarity and rank results.

We also conduct \textbf{human evaluation} by asking workers to rank the visualizations from all compared models in terms of their visual quality and character preservation.

\noindent\textbf{Visual Quality (Visual)} assess the overall image quality of a generated image sequence. In this aspect, we ask workers to focus on the image quality and check whether the objects and the background are clear.

\noindent\textbf{Character Preservation (Character)} checks the occurrences of characters. In this aspect, workers are requested to focus on the characters mentioned following the order of story sentences/paragraphs, and judge whether these characters appear in the image sequences.

\begin{table*}[]
\centering
\begin{tabular}{lccccc}
\toprule
Method        & Character F1 & Frame Accuracy & FID↓   & BLEU2/3   & R-Precision                            \\ \midrule
StoryGAN      & 18.59    & 9.34       & 158.06 & 3.24/1.22 & 1.51 ± 0.15                            \\
CP-CSV        & 21.78    & 10.03      & 149.29 & 3.25/1.22 & 1.76  ± 0.04                           \\ 
DUCO-StoryGAN & 38.01    & 13.97      & 96.51  & 3.68/1.34 & 3.56 ± 0.04                            \\ 
VLC-StoryGAN  & 43.02    & 17.36      & 84.96  & 3.80/1.44 & 3.28 ± 0.00                            \\ 
VQ-VAE-LM & 49.90    & 19.42      & 66.56 & 4.04/1.65 & 5.72± 0.02                        \\ 
\hdashline
 
+ Visual Planning  & 52.97    & 23.00      & 69.54 & 4.32/1.76 & 6.39 ± 0.00                            \\
+ Token Alignment  & 53.34    & 22.92      & \textbf{63.34} & 4.40/1.77 & 6.37 ± 0.00                         \\
\hdashline
VP-CSV   & \textbf{56.84}  & \textbf{25.87} &  65.51     &  \textbf{4.45}/\textbf{1.80}  &    \textbf{6.95 ± 0.00}             \\
 
\bottomrule


\end{tabular}
\caption{Automatic evaluation results on test split. All values are averaged over three runs. 
For the metrics, Character F1 and Frame Accuracy are associated with character preservation, FID is related to the image quality, while BLEU and R-Precision contribute to the semantic alignment.
We also conduct ablation studies on visual planning and character level token alignment. The results show that VP-CSV outperforms the previous GAN-based methods and baseline model VQ-VAE-LM in character preservation and semantic alignment. FID scores of previous works are obtained from \url{https://github.com/adymaharana/VLCStoryGan}. }
\label{tab:auto}
\vspace{-0.6cm}
\end{table*}

\subsection{Compared Methods}
We compare our proposed VP-CSV with the baseline VQ-VAE-LM and prior GAN-based models: 

\noindent\textbf{StoryGAN}~\cite{li2019storygan} uses the standard GAN technique, where a sequential model encodes a pair of story and image, and a story discriminator is trained with adversarial learning.

\noindent\textbf{CP-CSV}~\cite{song2020character} enhances its character preservation performance by including a character-level discriminator.

\noindent\textbf{DUCO-StoryGAN}~\cite{maharana2021improving} utilizes video captioning to conduct back-translation for adversarial learning and uses a transformer-based memory mechanism to encode the story.

\noindent \textbf{VLC-StoryGAN}~\cite{maharana2021integrating}, based on DUCO-StoryGAN, this model incorporates discourse information and an external knowledge source to enhance the visual quality and story-image consistency.


\subsection{Implementation Details}
For VQ-VAE training, all image data are scaled to $64 \times 64$ with patch size $=8$, resulting in a total visual token length $64/8 \times 64/8 \times 5 = 320$ per image sequence. The number of visual tokens is set to be $256$.
For the transformer, we use a 6-layer transformer model with dimension size of $768$.
We set the head number as $6$ and train a sparse transformer~\cite{child2019generating} with a dropout probability of $0.1$.
Exact architectural details and hyperparameters can be found in Appendix~\ref{sec:sup_implement}.

\section{Results}
Our experiments and analysis seek to answer these questions:
(1) Can VQ-VAE-LM outperform previous GAN-based models? If so, how?
(2) Whether the proposed VP-CSV improves the baseline VQ-VAE-LM model, especially in character preservation?
(3) How do the proposed two-stage visual planning and the character token alignment modules contribute to the performance improvements?

\paragraph{Main results.}
Table~\ref{tab:auto} summarizes the main results.
We can observe that the baseline VQ-VAE-LM outperforms the prior GAN-based methods by a large margin, and our proposed VP-CSV achieves even better results across all metrics.
Specifically, compared with the state-of-the-art GAN-based method, VLC-StoryGAN, our VP-CSV achieves 13.82\% (43.02\% $\rightarrow$ 56.84\%) and 8.51\% (17.36\% $\rightarrow$ 25.87\%) improvements per Character F1 score and Frame Accuracy, indicating stronger character preservation. Moreover, VP-CSV outperforms VLC-StoryGAN significantly per the back translation BLEU score and neural-based R-Precision score, showing better semantic alignment between generated images and story inputs.

\paragraph{Ablation study.}
Table~\ref{tab:auto} also shows ablation studies to ensure that each component in the our proposed method benefits story visualization.
We can observe that visual planning and token-level character alignment each improves character preservation scores (\ie Character F1, Frame Accuracy), resulting in better semantic alignment between generated images and the story (\ie BLEU2/3, R-Precision).
Combining them in VP-CSV achieves the best score on character preservation metrics.

\begin{table}[]
\small
\renewcommand*{\arraystretch}{1.4}
\centering
\begin{tabular}{|l|ccc|}

\hline
Metrics              & \multicolumn{3}{c|}{VLC. vs VQ-VAE-LM}                                                            \\\hline
               & \multicolumn{1}{c}{VLC.}        & \multicolumn{1}{c}{VQ-VAE-LM}        & \multicolumn{1}{c|}{Tie} \\ \cline{2-4}
\textbf{Visual} & 27.45                           & \textbf{62.75*} & 9.80                    \\
\textbf{Character}      & 37.25                           & \textbf{41.18}  & 21.57                   \\\hline
               & \multicolumn{3}{c|}{VQ-VAE-LM vs + TA}                                                            \\\cline{2-4}
               & \multicolumn{1}{c}{VQ-VAE-LM}       & \multicolumn{1}{c}{+ TA}         & \multicolumn{1}{c|}{Tie} \\\cline{2-4}
\textbf{Visual} & 33.33                           & \textbf{42.10}  & 24.56                   \\
\textbf{Character}      & 38.59                           & \textbf{40.35}  & 21.06                   \\\hline
               & \multicolumn{3}{c|}{VQ-VAE-LM vs VP-CSV}                                                          \\\cline{2-4}
               & \multicolumn{1}{c}{VQ-VAE-LM}       & \multicolumn{1}{c}{VP-CSV}       & \multicolumn{1}{c|}{Tie} \\\cline{2-4}
\textbf{Visual} &               34.51                &       \textbf{44.25}                         &         21.24           \\
\textbf{Character}      &            33.17                     &                \textbf{52.21*}                 &       14.62             \\\hline 
\end{tabular}
\caption{Pairwise human evaluation results. * denotes a significant different (p-value < 0.05) using sign test. We calculate the percentage of the human preference on the results from compared models. We see that VQ-VAE-LM can generate better image quality than GAN-based SOTA model (VLC.), while VP-CSV is better than VQ-VAE-LM at preserving characters.}
\label{tab:human}
\vspace{-0.3cm}
\end{table}

\paragraph{Human evaluation. } 

Since automatic metrics can be insufficient in evaluating the story visualization quality, we further conduct human evaluations by asking annotators to analyze the performance of the story visualization models. 
We hire in total 9 Amazon Mechanical Turkers who succeed in our previous large annotation tasks and passed the additional qualification exam for story visualization.
In total, we collect 218 sets of valid annotations story visualization results. 
By comparing models provided with the same input story, we ask annotators to rank the resulting image sequences. They rank the visualizations on a scale of 1 to 3, with 3 indicating the best and 1 otherwise.
Although ties are allowed, annotators are encouraged to provide unique rankings for each of the compared image sequences.
We show the detailed instructions and the user interface in Appendix~\ref{sec:instruction}.

Table~\ref{tab:human} shows the results of human evaluation. 
We report the pairwise results and run the sign test to show the significant difference. We can see that the VQ-VAE-LM baseline outperforms the previous works on visual quality and that VP-CSV also benefits from each proposed methods -- both visual planning and token-level character alignment improves the performance of character preservation.

\begin{figure}
    \centering
    \includegraphics[scale=0.37]{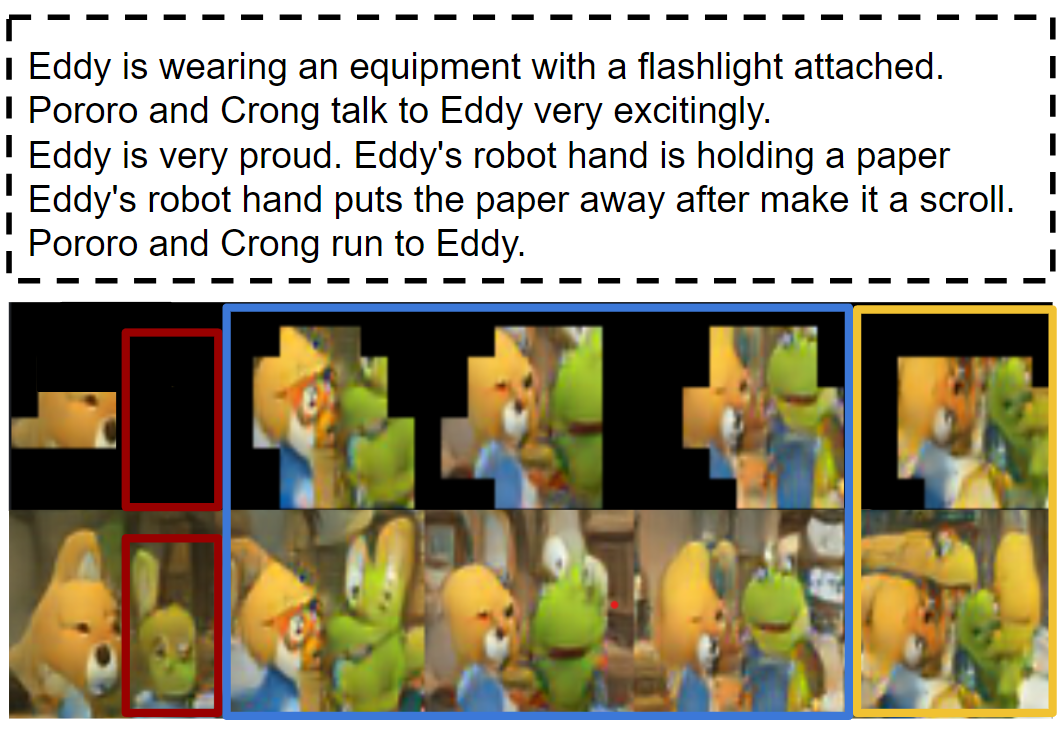}
    \caption{Case study on visual planning methods. Images in \blue{blue} boundary successfully follows our expectation of two-stage generation (character and background). Regions in \red{red} boundary shows a mistake that the \textbf{completion} model generates a character from no planned character token. The image in \orange{orange} boundary shows the low quality in characters.}
    \label{fig:analysis_vp}
    \vspace{-0.5cm}
\end{figure}

\paragraph{Analysis on visual planning.}
We conduct an in-depth study to analyze the performance of visual planning.
In Figure~\ref{fig:analysis_vp}, the first row shows the intermediate image results with the background masked, and the second row shows the final images.
We observe that our character token planning model successfully output character-related tokens in most of the cases (\blue{blue} boundary). This aligns with our expectation that our character-token plan module reinforces the system's attention to the appropriate characters. 
However, a mistake can e observed (\red{red} boundary) where the visual token completion model predicts a character based on nothing (masked background).
One possible explanation could be the discrepancy resulting from our two-stage modules being separately trained.
In addition, the fifth image (\orange{orange} boundary) reveals an example of low-quality character generation, which suggests future work to boost our model's visualization capacity.

\begin{figure}
    \centering
    \includegraphics[scale=0.2]{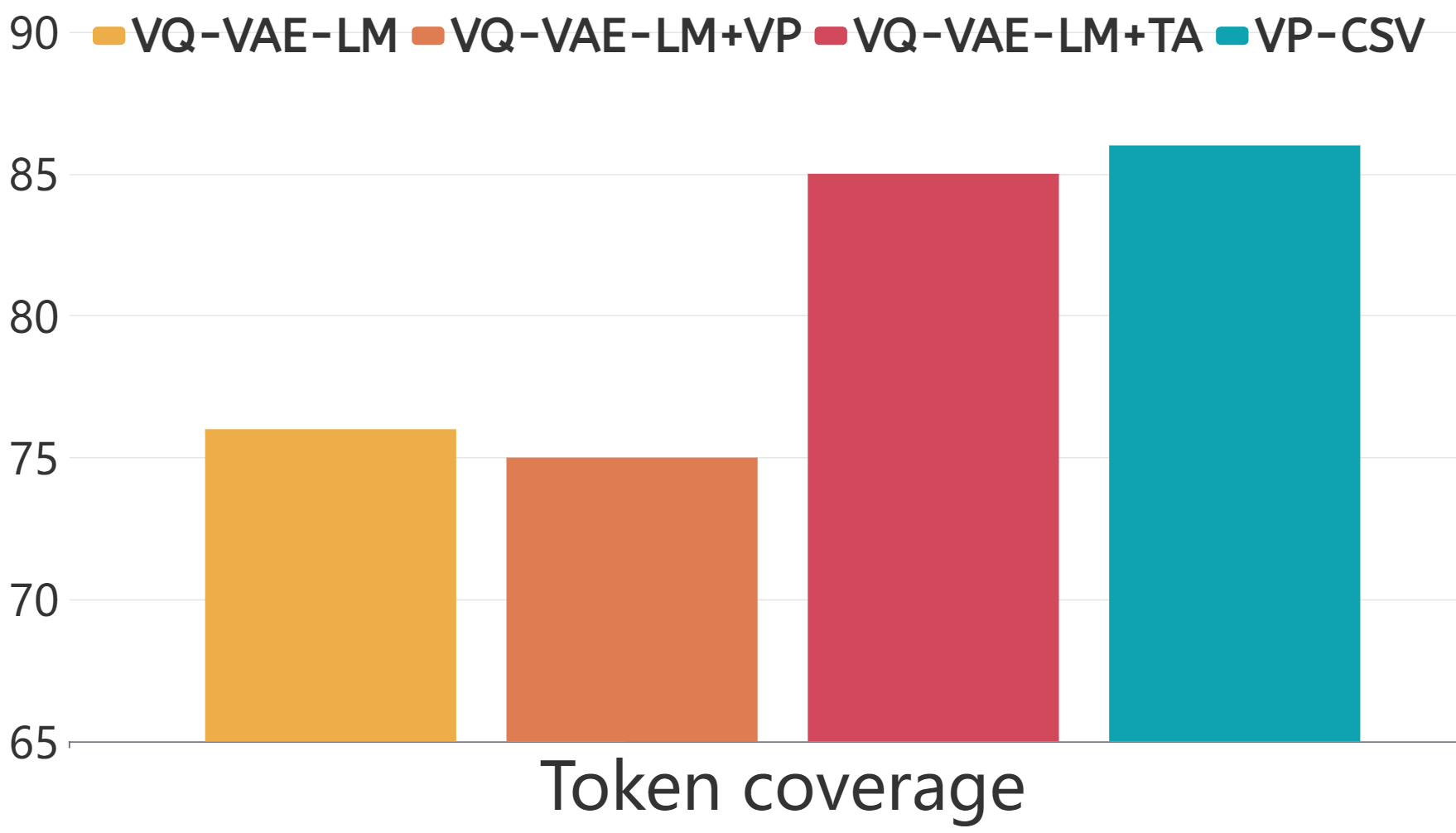}
    \caption{Character-token coverage ratio in the generated visual token sequences. Using character-token alignment (the last two columns) shows significant character token coverage improvement. }
    \label{fig:analysis_token}
    \vspace{-0.4cm}
\end{figure}

\paragraph{Analysis on character alignment.}
Our proposed character alignment method encourages the generated visual tokens to encompass the character-related tokens.
We calculated the percentage of character-token coverage in the generated visual token for the compared models.
Figure~\ref{fig:analysis_token} shows that our proposed VP-CSV model with character alignment can achieve the highest coverage, demonstrating the effectiveness of using semantic loss to preserve characters in the proposed framework.

\section{Conclusion} 
In this work, we propose visual planning and character token alignment to improve character preservation and visual quality. 
Extensive evaluation results show that our VP-CSV model outperforms the strong baseline models by a large margin. 
Future research in story visualization can aim at incorporating events that include not only characters but also their actions and relationships with one another, which furhter challenge machines' representation capability in story visualization.

\section{Limitations}
\label{sec:limitation}

\begin{figure}[h!]
    \centering
    \includegraphics[scale=0.6]{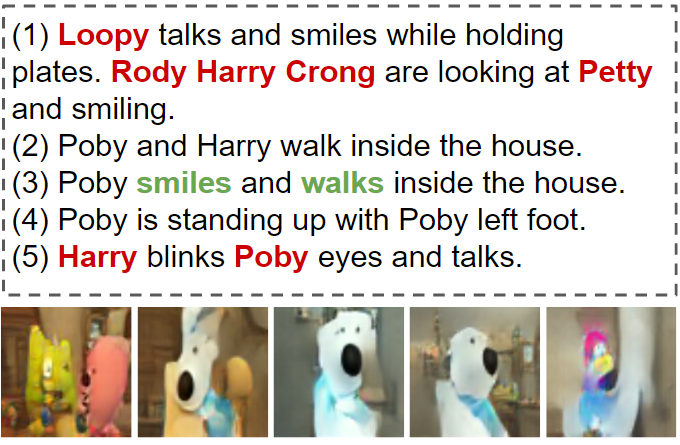}
    \caption{Limitation of generation image sequence.}
    \label{fig:limit}
\end{figure}

Our VP-CSV significantly improves the performance in story visualization, though there are still some limits as shown in Figure~\ref{fig:limit}. We summarize the limitations below.

\noindent \textbf{Multiple-character errors}. In the first image, we see an error when the model encounters multiple characters. With more characters in the image, it becomes harder to generate every individual in the image. In the fifth image, Harry and Poby are mixed, resulting in poor generation quality for both characters.

\noindent \textbf{Low image quality}. In the second image, Although Poby's nose is obvious, the rest of the body is vague. Improving overall image quality is certainly a future research direction.

\noindent \textbf{Handling Events}. In generated images, it is still hard to see the clear action performed by each character. For example, even though ``Poby'' is visible in the third image, it is uncertain if he is smiling or walking. It is challenging until the image quality is good enough to identify the event, so improving image quality could help identifying events in the future research.

\section{Ethics and Broader Impacts}

We hereby acknowledge that all of the co-authors of this work are aware of the provided \textit{ACM Code of Ethics} and honor the code of conduct.
This work is mainly about a two-stage module in story visualization task that follows a plan-and-write strategy.
The following gives the aspects of our ethical considerations and potential impact on the community.

\vspace{.3em}

\paragraph{Dataset.}
We collect the human annotation on evaluating the generated image sequences via Amazon Mechanical Turk (MTurk) and ensure that all the personal information of the workers involved (e.g., usernames, emails, URLs, demographic information, etc.) is discarded in the annotation.

The detailed annotation process (pay per amount of work, guidelines) is included in the appendix. Overall, we ensure our pay per task is above the annotator's local minimum wage (approximately \$15 USD / Hour).

\vspace{.3em}

\paragraph{Techniques.} 
As story evaluation is our main focus and we use an animation dataset as the testbed, we do not anticipate producing harmful outputs, especially towards vulnerable populations, by our proposed method.

\section*{Acknowledgments}
We thank PlusLab members at ULCA and USC for the helpful discussions, and the anonymous reviewers for useful feedback. This work was supported by DARPA Machine Common Sense (MCS) program under Cooperative Agreement N66001-19-2-4032, a Meta AI SRA, and JSPS/MEXT KAKENHI Grant Numbers JP19H04166 and JP22H05015. Part of the research results was obtained from the commissioned research (No. 225) by National Institute of Information and Communications Technology (NICT), Japan. Hong Chen was financially supported by Chair for Frontier AI Education, The University of Tokyo.

\bibliography{anthology,custom}

\begin{thebibliography}{37}
\expandafter\ifx\csname natexlab\endcsname\relax\def\natexlab#1{#1}\fi

\bibitem[{Chattopadhay et~al.(2018)Chattopadhay, Sarkar, Howlader, and
  Balasubramanian}]{chattopadhay2018grad}
Aditya Chattopadhay, Anirban Sarkar, Prantik Howlader, and Vineeth~N
  Balasubramanian. 2018.
\newblock Grad-cam++: Generalized gradient-based visual explanations for deep
  convolutional networks.
\newblock In \emph{2018 IEEE winter conference on applications of computer
  vision (WACV)}, pages 839--847. IEEE.

\bibitem[{Child et~al.(2019)Child, Gray, Radford, and
  Sutskever}]{child2019generating}
Rewon Child, Scott Gray, Alec Radford, and Ilya Sutskever. 2019.
\newblock Generating long sequences with sparse transformers.
\newblock \emph{arXiv preprint arXiv:1904.10509}.

\bibitem[{Ding et~al.(2021)Ding, Yang, Hong, Zheng, Zhou, Yin, Lin, Zou, Shao,
  Yang et~al.}]{ding2021cogview}
Ming Ding, Zhuoyi Yang, Wenyi Hong, Wendi Zheng, Chang Zhou, Da~Yin, Junyang
  Lin, Xu~Zou, Zhou Shao, Hongxia Yang, et~al. 2021.
\newblock Cogview: Mastering text-to-image generation via transformers.
\newblock \emph{Advances in Neural Information Processing Systems},
  34:19822--19835.

\bibitem[{Fu et~al.(2020)Fu, Hu, Dong, Guo, Gao, and Li}]{fu2020axiom}
Ruigang Fu, Qingyong Hu, Xiaohu Dong, Yulan Guo, Yinghui Gao, and Biao Li.
  2020.
\newblock Axiom-based grad-cam: Towards accurate visualization and explanation
  of cnns.
\newblock \emph{arXiv preprint arXiv:2008.02312}.

\bibitem[{Goldfarb-Tarrant et~al.(2020)Goldfarb-Tarrant, Chakrabarty,
  Weischedel, and Peng}]{goldfarb-tarrant-etal-2020-content}
Seraphina Goldfarb-Tarrant, Tuhin Chakrabarty, Ralph Weischedel, and Nanyun
  Peng. 2020.
\newblock \href {https://doi.org/10.18653/v1/2020.emnlp-main.351} {Content
  planning for neural story generation with aristotelian rescoring}.
\newblock In \emph{Proceedings of the 2020 Conference on Empirical Methods in
  Natural Language Processing (EMNLP)}, pages 4319--4338, Online. Association
  for Computational Linguistics.

\bibitem[{Goldfarb-Tarrant et~al.(2019)Goldfarb-Tarrant, Feng, and
  Peng}]{goldfarb2019plan}
Seraphina Goldfarb-Tarrant, Haining Feng, and Nanyun Peng. 2019.
\newblock Plan, write, and revise: an interactive system for open-domain story
  generation.
\newblock In \emph{2019 Annual Conference of the North American Chapter of the
  Association for Computational Linguistics (NAACL-HLT 2019), Demonstrations
  Track}, volume~4, pages 89--97.

\bibitem[{Goodfellow et~al.(2014)Goodfellow, Pouget-Abadie, Mirza, Xu,
  Warde-Farley, Ozair, Courville, and Bengio}]{goodfellow2014generative}
Ian Goodfellow, Jean Pouget-Abadie, Mehdi Mirza, Bing Xu, David Warde-Farley,
  Sherjil Ozair, Aaron Courville, and Yoshua Bengio. 2014.
\newblock Generative adversarial nets.
\newblock \emph{Advances in neural information processing systems}, 27.

\bibitem[{Gupta et~al.(2018)Gupta, Schwenk, Farhadi, Hoiem, and
  Kembhavi}]{gupta2018imagine}
Tanmay Gupta, Dustin Schwenk, Ali Farhadi, Derek Hoiem, and Aniruddha Kembhavi.
  2018.
\newblock Imagine this! scripts to compositions to videos.
\newblock In \emph{Proceedings of the European Conference on Computer Vision
  (ECCV)}, pages 598--613.

\bibitem[{Han et~al.(2022)Han, Chen, Tian, and Peng}]{han2022go}
Rujun Han, Hong Chen, Yufei Tian, and Nanyun Peng. 2022.
\newblock Go back in time: Generating flashbacks in stories with event temporal
  prompts.
\newblock In \emph{2022 Annual Conference of the North American Chapter of the
  Association for Computational Linguistics (NAACL)}.

\bibitem[{Kingma and Welling(2014)}]{kingma2014auto}
Diederik~P Kingma and Max Welling. 2014.
\newblock Auto-encoding variational bayes.
\newblock \emph{stat}, 1050:10.

\bibitem[{Kodali et~al.(2017)Kodali, Abernethy, Hays, and
  Kira}]{kodali2017convergence}
Naveen Kodali, Jacob Abernethy, James Hays, and Zsolt Kira. 2017.
\newblock On convergence and stability of gans.
\newblock \emph{arXiv preprint arXiv:1705.07215}.

\bibitem[{Li et~al.(2020)Li, Kong, and Zhou}]{li2020improved}
Chunye Li, Liya Kong, and Zhiping Zhou. 2020.
\newblock Improved-storygan for sequential images visualization.
\newblock \emph{Journal of Visual Communication and Image Representation},
  73:102956.

\bibitem[{Li et~al.(2019)Li, Gan, Shen, Liu, Cheng, Wu, Carin, Carlson, and
  Gao}]{li2019storygan}
Yitong Li, Zhe Gan, Yelong Shen, Jingjing Liu, Yu~Cheng, Yuexin Wu, Lawrence
  Carin, David Carlson, and Jianfeng Gao. 2019.
\newblock Storygan: A sequential conditional gan for story visualization.
\newblock In \emph{Proceedings of the IEEE/CVF Conference on Computer Vision
  and Pattern Recognition}, pages 6329--6338.

\bibitem[{Maharana and Bansal(2021)}]{maharana2021integrating}
Adyasha Maharana and Mohit Bansal. 2021.
\newblock Integrating visuospatial, linguistic, and commonsense structure into
  story visualization.
\newblock In \emph{Proceedings of the 2021 Conference on Empirical Methods in
  Natural Language Processing}, pages 6772--6786.

\bibitem[{Maharana et~al.(2021)Maharana, Hannan, and
  Bansal}]{maharana2021improving}
Adyasha Maharana, Darryl Hannan, and Mohit Bansal. 2021.
\newblock Improving generation and evaluation of visual stories via semantic
  consistency.
\newblock In \emph{Proceedings of the 2021 Conference of the North American
  Chapter of the Association for Computational Linguistics: Human Language
  Technologies}, pages 2427--2442.

\bibitem[{Mirza and Osindero(2014)}]{mirza2014conditional}
Mehdi Mirza and Simon Osindero. 2014.
\newblock Conditional generative adversarial nets.
\newblock \emph{arXiv preprint arXiv:1411.1784}.

\bibitem[{Montgomery(2004)}]{montgomery2004language}
HEMINGWAY SHORT STORY~Martin Montgomery. 2004.
\newblock Language, character and action: A linguistic approach to the analysis
  of character in a hemingway short story.
\newblock In \emph{Techniques of Description}, pages 143--158. Routledge.

\bibitem[{Muhammad and Yeasin(2020)}]{muhammad2020eigen}
Mohammed~Bany Muhammad and Mohammed Yeasin. 2020.
\newblock Eigen-cam: Class activation map using principal components.
\newblock \emph{arXiv preprint arXiv:2008.00299}.

\bibitem[{Peng et~al.(2018)Peng, Ghazvininejad, May, and
  Knight}]{peng2018towards}
Nanyun Peng, Marjan Ghazvininejad, Jonathan May, and Kevin Knight. 2018.
\newblock Towards controllable story generation.
\newblock In \emph{NAACL Workshop}.

\bibitem[{Qiao et~al.(2019)Qiao, Zhang, Xu, and Tao}]{qiao2019mirrorgan}
Tingting Qiao, Jing Zhang, Duanqing Xu, and Dacheng Tao. 2019.
\newblock Mirrorgan: Learning text-to-image generation by redescription.
\newblock In \emph{Proceedings of the IEEE/CVF Conference on Computer Vision
  and Pattern Recognition}, pages 1505--1514.

\bibitem[{Radford et~al.(2019)Radford, Wu, Child, Luan, Amodei, and
  Sutskever}]{radford2019language}
Alec Radford, Jeff Wu, Rewon Child, David Luan, Dario Amodei, and Ilya
  Sutskever. 2019.
\newblock Language models are unsupervised multitask learners.

\bibitem[{Ramaswamy et~al.(2020)}]{ramaswamy2020ablation}
Harish~Guruprasad Ramaswamy et~al. 2020.
\newblock Ablation-cam: Visual explanations for deep convolutional network via
  gradient-free localization.
\newblock In \emph{Proceedings of the IEEE/CVF Winter Conference on
  Applications of Computer Vision}, pages 983--991.

\bibitem[{Ramesh et~al.(2021)Ramesh, Pavlov, Goh, Gray, Voss, Radford, Chen,
  and Sutskever}]{ramesh2021zero}
Aditya Ramesh, Mikhail Pavlov, Gabriel Goh, Scott Gray, Chelsea Voss, Alec
  Radford, Mark Chen, and Ilya Sutskever. 2021.
\newblock Zero-shot text-to-image generation.
\newblock In \emph{International Conference on Machine Learning}, pages
  8821--8831. PMLR.

\bibitem[{Selvaraju et~al.(2017)Selvaraju, Cogswell, Das, Vedantam, Parikh, and
  Batra}]{selvaraju2017grad}
Ramprasaath~R Selvaraju, Michael Cogswell, Abhishek Das, Ramakrishna Vedantam,
  Devi Parikh, and Dhruv Batra. 2017.
\newblock Grad-cam: Visual explanations from deep networks via gradient-based
  localization.
\newblock In \emph{Proceedings of the IEEE international conference on computer
  vision}, pages 618--626.

\bibitem[{Sohn et~al.(2015)Sohn, Lee, and Yan}]{sohn2015learning}
Kihyuk Sohn, Honglak Lee, and Xinchen Yan. 2015.
\newblock Learning structured output representation using deep conditional
  generative models.
\newblock \emph{Advances in neural information processing systems}, 28.

\bibitem[{Song et~al.(2020)Song, Rui~Tam, Chen, Lu, and
  Shuai}]{song2020character}
Yun-Zhu Song, Zhi Rui~Tam, Hung-Jen Chen, Huiao-Han Lu, and Hong-Han Shuai.
  2020.
\newblock Character-preserving coherent story visualization.
\newblock In \emph{European Conference on Computer Vision}, pages 18--33.
  Springer.

\bibitem[{Szegedy et~al.(2016)Szegedy, Vanhoucke, Ioffe, Shlens, and
  Wojna}]{szegedy2016rethinking}
Christian Szegedy, Vincent Vanhoucke, Sergey Ioffe, Jon Shlens, and Zbigniew
  Wojna. 2016.
\newblock Rethinking the inception architecture for computer vision.
\newblock In \emph{Proceedings of the IEEE conference on computer vision and
  pattern recognition}, pages 2818--2826.

\bibitem[{Thanh-Tung et~al.(2018)Thanh-Tung, Tran, and
  Venkatesh}]{thanh2018improving}
Hoang Thanh-Tung, Truyen Tran, and Svetha Venkatesh. 2018.
\newblock Improving generalization and stability of generative adversarial
  networks.
\newblock In \emph{International Conference on Learning Representations}.

\bibitem[{Van Den~Oord et~al.(2017)Van Den~Oord, Vinyals
  et~al.}]{van2017neural}
Aaron Van Den~Oord, Oriol Vinyals, et~al. 2017.
\newblock Neural discrete representation learning.
\newblock \emph{Advances in neural information processing systems}, 30.

\bibitem[{Wang et~al.(2020)Wang, Wang, Du, Yang, Zhang, Ding, Mardziel, and
  Hu}]{wang2020score}
Haofan Wang, Zifan Wang, Mengnan Du, Fan Yang, Zijian Zhang, Sirui Ding, Piotr
  Mardziel, and Xia Hu. 2020.
\newblock Score-cam: Score-weighted visual explanations for convolutional
  neural networks.
\newblock In \emph{Proceedings of the IEEE/CVF conference on computer vision
  and pattern recognition workshops}, pages 24--25.

\bibitem[{Xu et~al.(2018{\natexlab{a}})Xu, Zhang, Friedman, Liang, and
  Broeck}]{xu2018semantic}
Jingyi Xu, Zilu Zhang, Tal Friedman, Yitao Liang, and Guy Broeck.
  2018{\natexlab{a}}.
\newblock A semantic loss function for deep learning with symbolic knowledge.
\newblock In \emph{International conference on machine learning}, pages
  5502--5511. PMLR.

\bibitem[{Xu et~al.(2018{\natexlab{b}})Xu, Zhang, Huang, Zhang, Gan, Huang, and
  He}]{xu2018attngan}
Tao Xu, Pengchuan Zhang, Qiuyuan Huang, Han Zhang, Zhe Gan, Xiaolei Huang, and
  Xiaodong He. 2018{\natexlab{b}}.
\newblock Attngan: Fine-grained text to image generation with attentional
  generative adversarial networks.
\newblock In \emph{Proceedings of the IEEE conference on computer vision and
  pattern recognition}, pages 1316--1324.

\bibitem[{Yan et~al.(2021)Yan, Zhang, Abbeel, and Srinivas}]{yan2021videogpt}
Wilson Yan, Yunzhi Zhang, Pieter Abbeel, and Aravind Srinivas. 2021.
\newblock Videogpt: Video generation using vq-vae and transformers.
\newblock \emph{arXiv preprint arXiv:2104.10157}.

\bibitem[{Yao et~al.(2019)Yao, Peng, Weischedel, Knight, Zhao, and
  Yan}]{yao2019plan}
Lili Yao, Nanyun Peng, Ralph Weischedel, Kevin Knight, Dongyan Zhao, and Rui
  Yan. 2019.
\newblock Plan-and-write: Towards better automatic storytelling.
\newblock In \emph{Proceedings of the AAAI Conference on Artificial
  Intelligence}, volume~33, pages 7378--7385.

\bibitem[{Yu et~al.(2019)Yu, Zhang, Cao, and Xia}]{yu2019vaegan}
Xianwen Yu, Xiaoning Zhang, Yang Cao, and Min Xia. 2019.
\newblock Vaegan: A collaborative filtering framework based on adversarial
  variational autoencoders.
\newblock In \emph{IJCAI}, pages 4206--4212.

\bibitem[{Zeng et~al.(2019)Zeng, Li, and Zhang}]{zeng2019pororogan}
Gangyan Zeng, Zhaohui Li, and Yuan Zhang. 2019.
\newblock Pororogan: An improved story visualization model on pororo-sv
  dataset.
\newblock In \emph{Proceedings of the 2019 3rd International Conference on
  Computer Science and Artificial Intelligence}, pages 155--159.

\bibitem[{Zhang et~al.(2017)Zhang, Xu, Li, Zhang, Wang, Huang, and
  Metaxas}]{zhang2017stackgan}
Han Zhang, Tao Xu, Hongsheng Li, Shaoting Zhang, Xiaogang Wang, Xiaolei Huang,
  and Dimitris~N Metaxas. 2017.
\newblock Stackgan: Text to photo-realistic image synthesis with stacked
  generative adversarial networks.
\newblock In \emph{Proceedings of the IEEE international conference on computer
  vision}, pages 5907--5915.

\end{thebibliography}
\bibliographystyle{acl_natbib}
\clearpage
\appendix

\section{Appendix}
\label{sec:appendix}

\subsection{More examples}
\begin{figure}[h]
    \centering
    \includegraphics[scale=0.4]{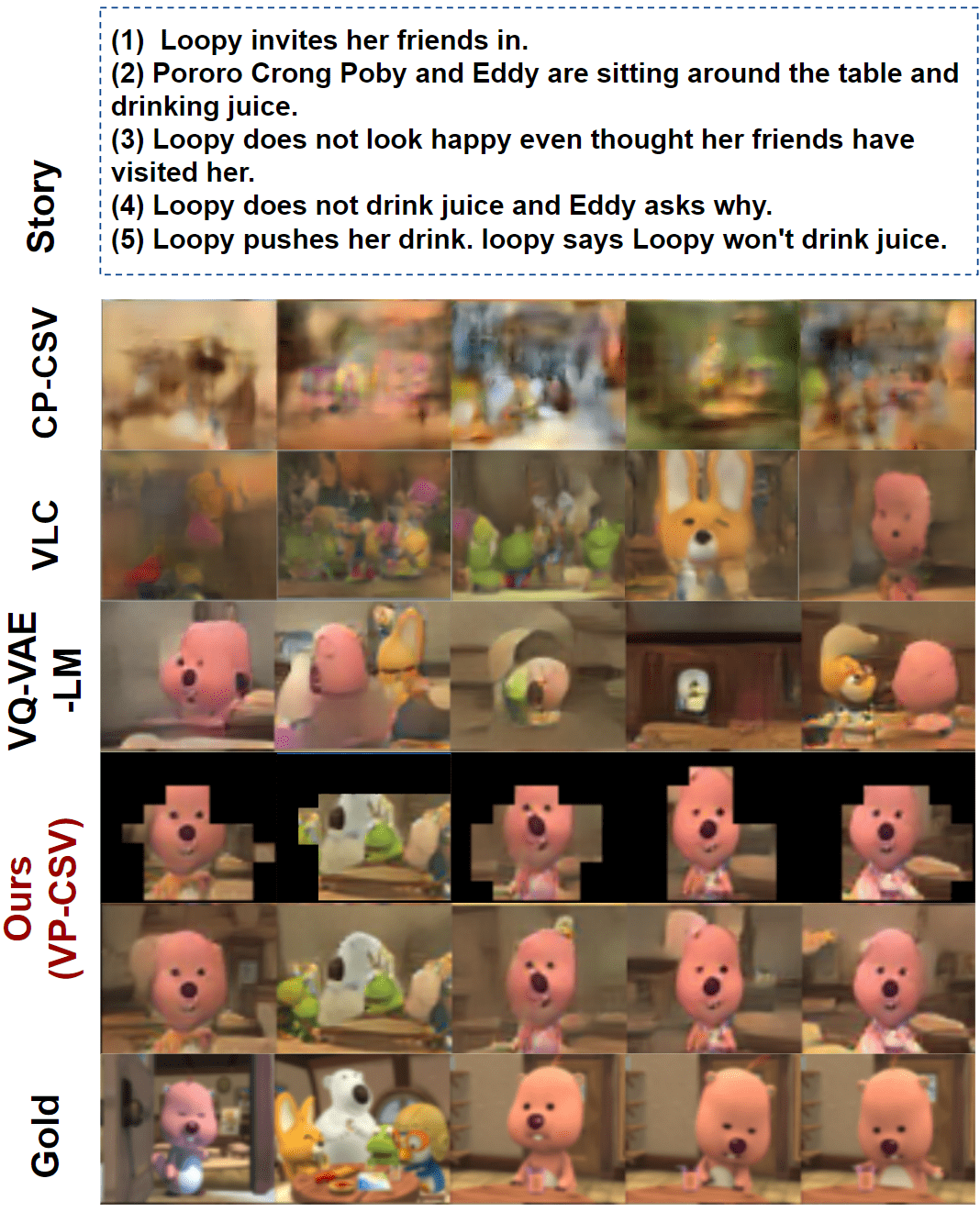}
    \caption{This example shows that our VP-CSV exactly follows the given stories, and can generate similar characters compared with the gold image sequence.}
    \label{fig:supdemo1}
\end{figure}

\begin{figure}
    \centering
    \includegraphics[scale=0.4]{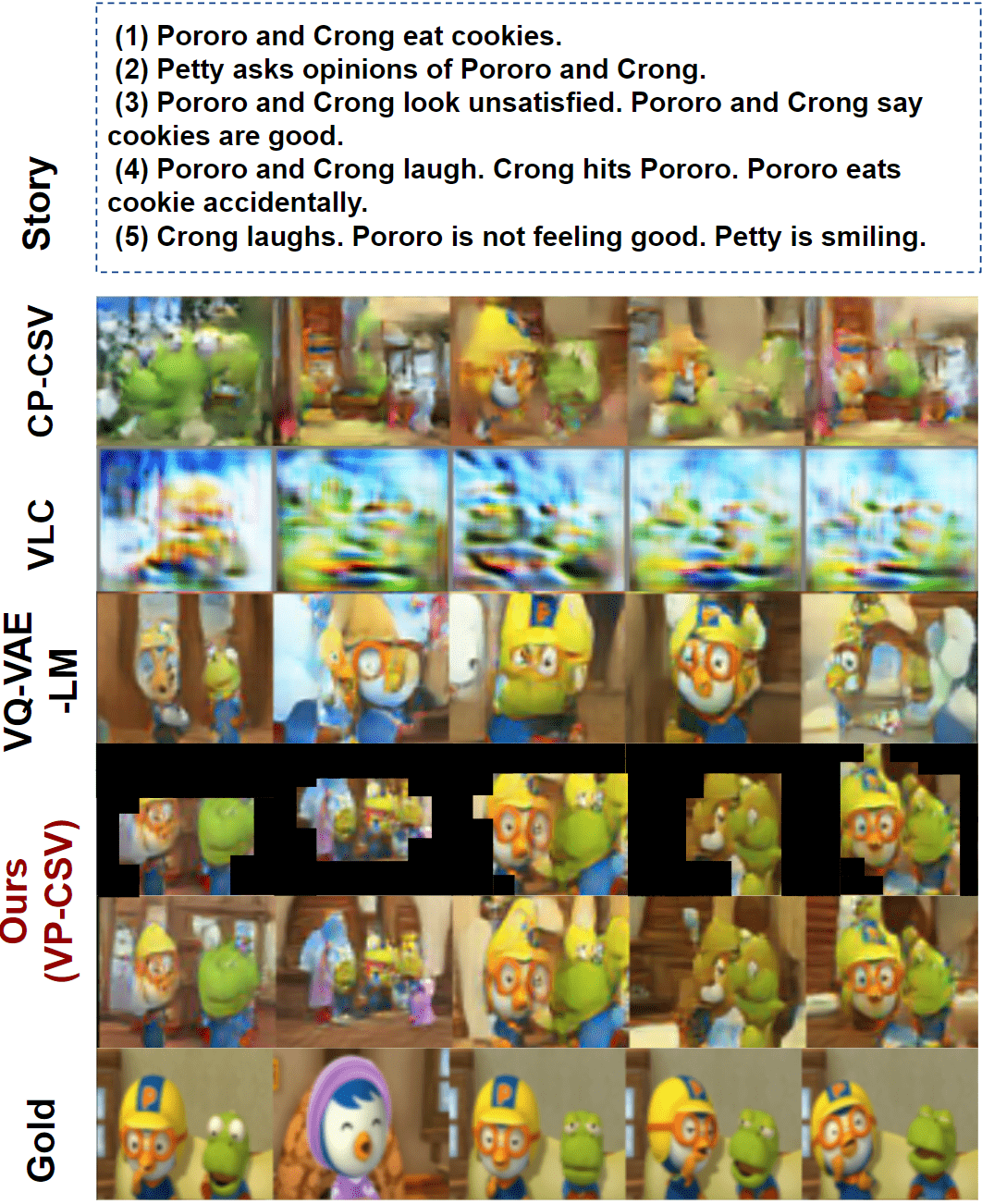}
    \caption{This example shows that GAN-based models tends to generate vague and incorrect images (especially VLC in this case), indicating the instability of GAN-based models.}
    \label{fig:supdemo2}
\end{figure}

We show another two examples of the results generated by different methods.
In Figure~\ref{fig:supdemo1}, we observe that the image sequence generated by our VP-CSV precisely matches the given story. The characters are likewise compatible with the gold image sequence; however, the characters are unclear in the second image. In Figure~\ref{fig:supdemo2}, compared with baseline model VQ-VAE-LM, VP-CSV generates better `Pororo'' and ``Crong''. Besides, the results of GAN-based approaches (i.e. VLC and CP-CSV) are exceedingly imprecise, making it almost impossible to distinguish the characters in the image sequence.

\begin{figure}[t]
    \centering
    \includegraphics[scale=0.4]{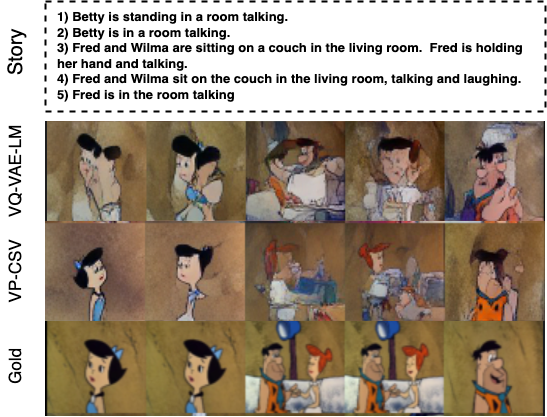}
    \caption{Examples on Flintstones dataset.}
    \label{fig:flintstone}
\end{figure}

\paragraph{Results on Flintstones dataset.}
FlintstonesSV~\cite{maharana2021integrating} is another story visualization dataset that is derived by Flintstones dataset~\cite{gupta2018imagine}.
To show the generalization of our proposed VP-CSV method, we provide an example on Flintstones dataset in Figure~\ref{fig:flintstone}. We can observe that compared with VQ-VAE-LM, our proposed method provides higher character quality.

\subsection{Grad-CAM results}
\label{sec:gradcam_app}
\begin{figure}
    \centering
    \includegraphics[scale=0.18]{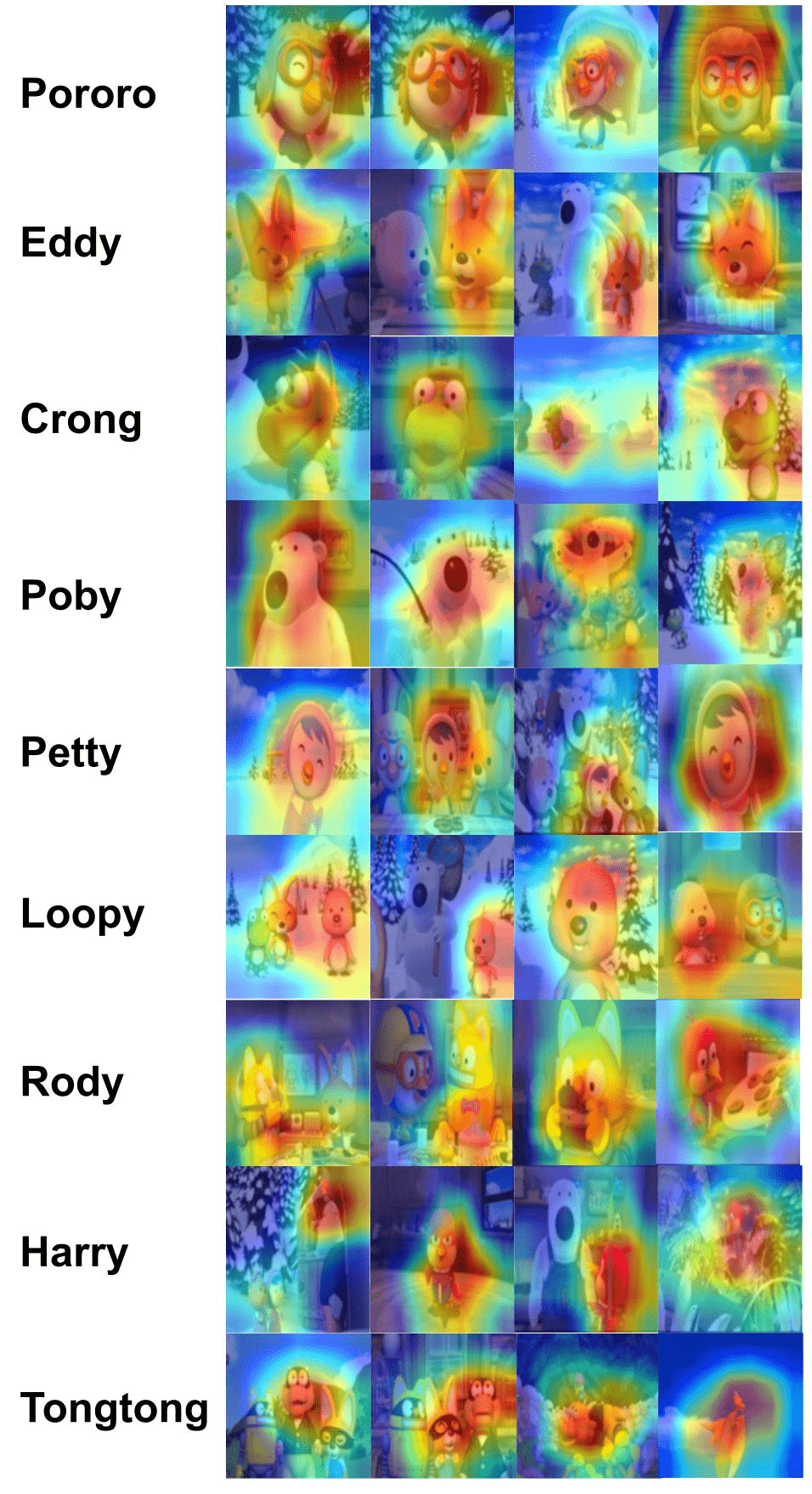}
    \caption{Grad-CAM examples for each character. We separately show the visualization results for each character. We can see that Grad-CAM can correctly locate the character regions in most of the cases.}
    \label{fig:sup_gradcam}
\end{figure}

In Figure~\ref{fig:sup_gradcam}, we show some Grad-CAM results on each character. We can see that most of the character regions are highlighted, indicating the high performance of our character region extraction.

\paragraph{Analysis on Grad-CAM variants.}
Since many recent works have extended the original Gran-CAM, we compare the results of different CAM methods, including  XGrad-CAM~\cite{fu2020axiom}, Grad-CAM++~\cite{chattopadhay2018grad}, Eigen-CAM~\cite{muhammad2020eigen}, Ablation-CAM~\cite{ramaswamy2020ablation} and Score-CAM~\cite{wang2020score}. In Table~\ref{tab:gradcam_variant}, we can see that almost all the CAM methods (except Eigen-CAM) can capture the characters' regions. However, in terms of the inference time, Grad-CAM is the fastest as it is the most straightforward method.
Therefore, we use Grad-CAM in our experiment for time-saving character region extraction.

\begin{table}

\centering
\includegraphics[scale=0.26]{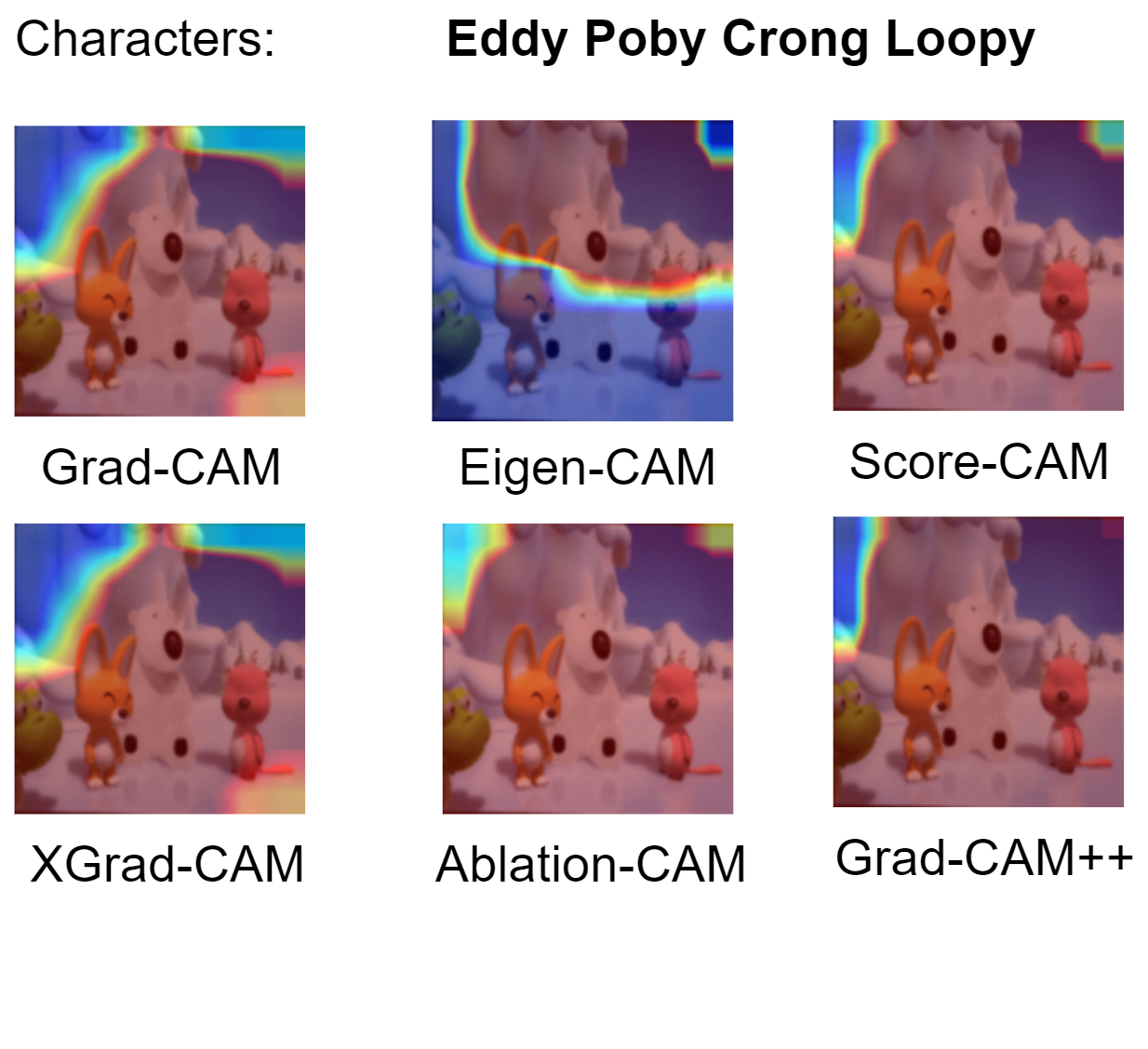}\\

    \begin{tabular}{lc}
    \toprule
    ResNet-50 & \makecell{Inference Time \\(sec/image)}\\\midrule
    Grad-CAM & 0.1906\\
    XGrad-CAM & 0.2026\\
    Grad-CAM++ & 0.2074\\
    Eigen-CAM & 1.2671\\
    Ablation-CAM & 10.0934\\
    Score-CAM & 12.5859\\\bottomrule
    \end{tabular}
\caption{Examples generated by variants of Grad-CAM and their inference time.}
\label{tab:gradcam_variant}
\end{table}

\begin{figure}[!htbp]
    \begin{minipage}{0.45\textwidth}
        \centering
        \includegraphics[scale=0.4]{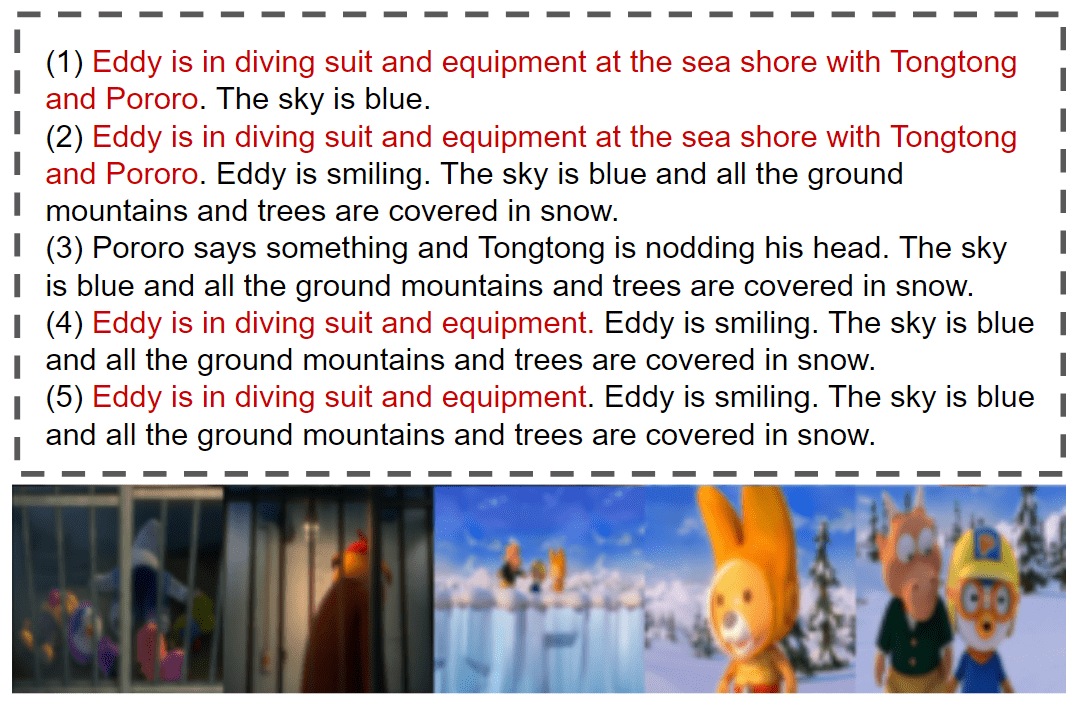}
        \caption{Repetition in story sentences.}
        \label{fig:dataset_limit1}
    \end{minipage}\hfill
    \begin{minipage}{0.45\textwidth}
        \centering
        \includegraphics[scale=0.4]{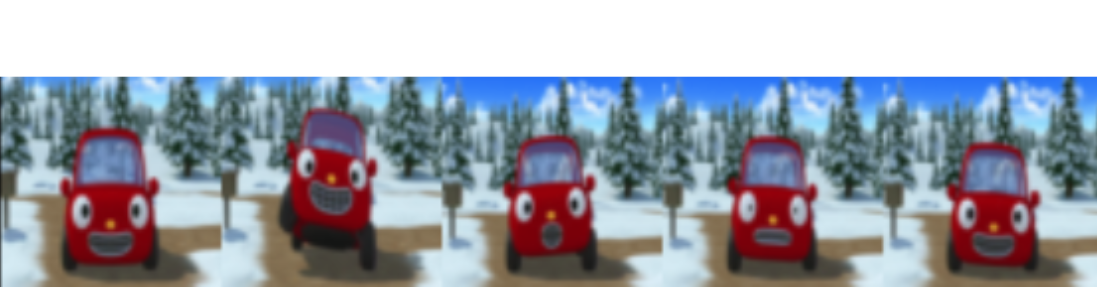}
        \caption{Repetition in images.}
        \label{fig:dataset_limit2}
    \end{minipage}
\end{figure}

\subsection{Limitations on Pororo-SV dataset}
In this section, we discuss the limitations of the Pororo-SV dataset. Figure~\ref{fig:dataset_limit1} and Figure~\ref{fig:dataset_limit2} show bad training samples in the Pororo-SV dataset. Figure~\ref{fig:dataset_limit1} shows serious repetition issues (in \red{red}) in some of the input stories. Obviously, concatenating these short paragraphs cannot compose a story, thus deviating from purpose of story visualization task. 
It will cause our model to learn lousy alignment between the story and the image sequence.
Besides, repetition issues can also happen on the image side. As shown in Figure~\ref{fig:dataset_limit2}. Due to the frame extraction problem, sometimes the image sequence in the dataset is almost in one scene with no significant changes. This will cause our model to output similar images in the resulted image sequence. 

\subsection{Train-Test discrepancy}
In this paper, we propose using VQ-VAE-LM and adapting its Transformer into a two-stage model. We encounter two train-test discrepancy issues: independent training of VQ-VAE and Transformer and the independent training of two stages. Due to the extraordinarily high time-consuming joint training of these models, we place them into our future works.

\subsection{Implementation details}
\label{sec:sup_implement}
We scale the image data into 64×64 as the same as previous works~\cite{li2019storygan, maharana2021integrating}.
Our VQ-VAE is pretrained on a single image in the Pororo training dataset. Each image can be quantized into an 8×8 visual token matrix, where each visual token represents a region of 8×8 pixels. We use one NVIDIA A100 40GB to train this model for 48 hours. 
For training transformer, both character token planning and visual token completion models use a 6-layer transformer model with dimension size of 768. We set the head number as 6 and train Sparse Transformers~\cite{child2019generating} with local and strided attention across space-time with a dropout probability of 0.1.
We use four NVIDIA A100 40GB to train each model for 12 hours. The implementations of the transformer-based models are extended from the DALLE-python code~\footnote{\url{https://github.com/lucidrains/DALLE-pytorch}},
and our entire code-base is implemented in PyTorch~\footnote{\url{https://pytorch.org/}}.
Please note that all the models in this paper are trained from scratch.

\paragraph{Hyperparameters}
The results are averaged over three runs until performance convergence is observed.
We tame the image patch size in [4,8,12,16], the transformer dimension in [512, 768, 1024], the number of visual tokens in [64, 128, 256, 512], and choose the hyperparameters with the best score on FID (\ie image quality).

\subsection{Annotation instruction}
\label{sec:instruction}
\begin{figure}[!t]
    \centering
    \includegraphics[scale=0.65]{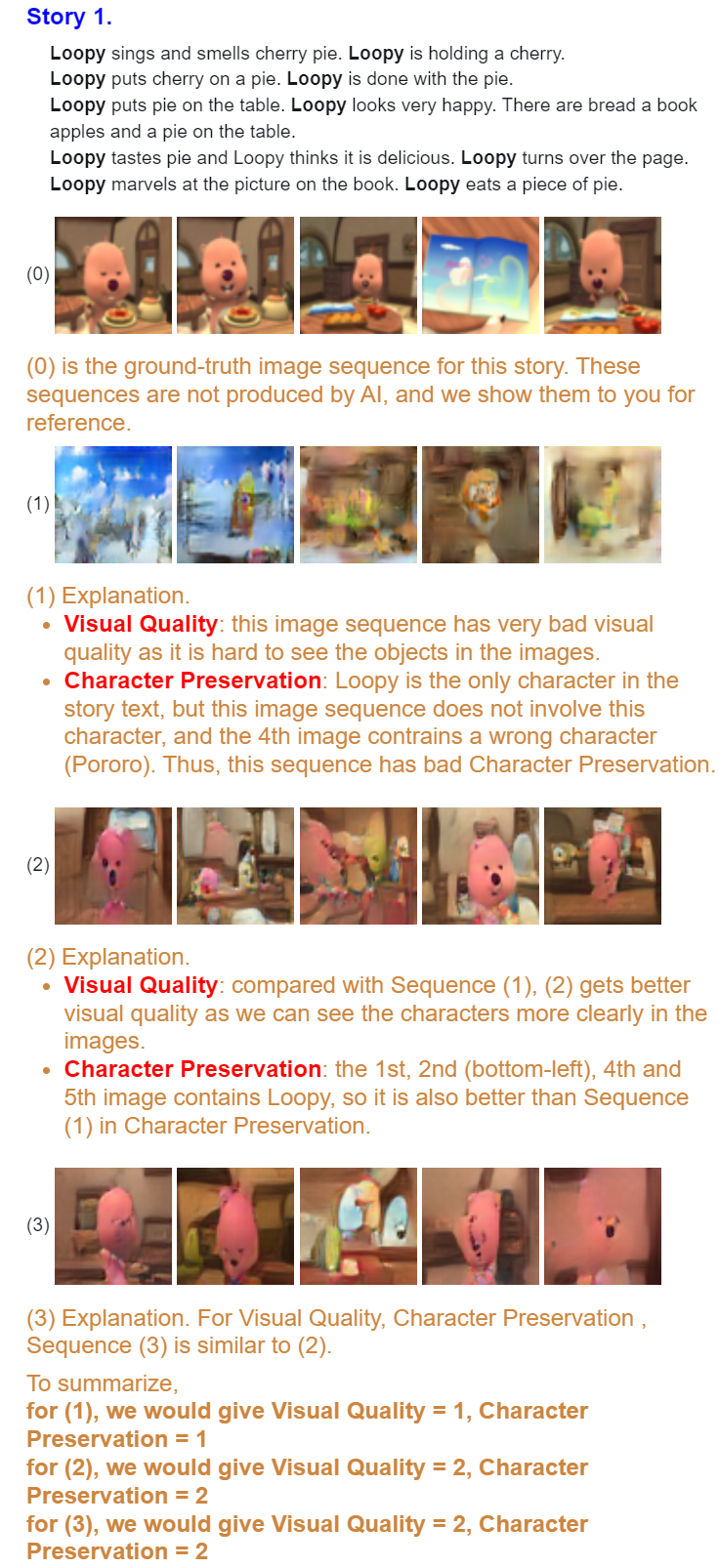}
    \caption{Example of our instruction.}
    \label{fig:instruction1}
\end{figure}

We request AMT for human evaluation. We first launch the qualification round to determine the annotators' qualifications. Each annotator can submit up to 10 HITs (\$1.5 for each annotation). We provide detailed instructions to guide annotation. Finally, we send \$8 bonus for qualified annotators and invite them for the next rounds. The instruction is shown in Figure~\ref{fig:instruction1}.

\end{document}